%% file: main.tex
\RequirePackage{fix-cm}
\documentclass[11pt]{article}

\usepackage[preprint]{acl}

\usepackage{fontspec}
\usepackage{amsmath,amssymb,bm}
\usepackage{booktabs,tabularx,array,multirow}
\usepackage{graphicx}
\usepackage{xcolor}
\usepackage{float}
\usepackage{microtype}
\usepackage[nameinlink,noabbrev]{cleveref}
\crefname{figure}{Figure}{Figures}
\crefname{table}{Table}{Tables}
\usepackage{enumitem}
\usepackage{tikz}
\usetikzlibrary{positioning,arrows.meta,fit,backgrounds,calc}

\graphicspath{{figures/}}
\setlist{nosep,leftmargin=*}
\renewcommand{\arraystretch}{1.05}

\makeatletter
\newenvironment{papertable}{%
  \if@twocolumn\begin{table*}[!t]\else\begin{table}[H]\fi
}{%
  \if@twocolumn\end{table*}\else\end{table}\fi
}
\newenvironment{paperfigure}{%
  \if@twocolumn\begin{figure*}[!t]\else\begin{figure}[H]\fi
}{%
  \if@twocolumn\end{figure*}\else\end{figure}\fi
}
\makeatother

\newcommand{\Sstar}{S^{\ast}}
\newcommand{\deltah}{\Delta_H}
\newcommand{\pathA}{\textnormal{A}}
\newcommand{\pathB}{\textnormal{B}}
\newcommand{\pathC}{\textnormal{C}}
\newcommand{\code}[1]{\texttt{#1}}

\makeatletter
\newcommand{\appendixnotes}{}
\newcommand{\appendixnote}[2]{%
  \g@addto@macro\appendixnotes{\appendixnoteentry{#1}{#2}}%
  \phantomsection\label{#1}\appendixnoteref{#1}}
\newcommand{\appendixnoteref}[1]{\textsuperscript{\hyperlink{#1-note}{*}}}
\newcommand{\appendixnoteentry}[2]{%
  \ifnum\getpagerefnumber{#1}=\value{page}\relax
    \expandafter\def\expandafter\appendixpagenotes\expandafter{%
      \appendixpagenotes\par\noindent\hypertarget{#1-note}{*}~#2}%
  \fi}
\AddToHook{shipout/foreground}{%
  \begingroup
  \let\appendixpagenotes\@empty
  \appendixnotes
  \ifx\appendixpagenotes\@empty\else
    \begin{tikzpicture}[remember picture,overlay]
      \node[anchor=south west,inner sep=0pt,xshift=12mm,yshift=12mm]
        at (current page.south west) {%
          \begin{minipage}{\dimexpr\paperwidth-24mm\relax}
            \footnotesize\linespread{1}\selectfont\appendixpagenotes
          \end{minipage}};
    \end{tikzpicture}%
  \fi
  \endgroup}
\makeatother

\newcommand{\figslot}[2]{%
  \IfFileExists{figures/#1.pdf}{\includegraphics[width=\linewidth]{#1}}{%
    \IfFileExists{figures/#1.png}{\includegraphics[width=\linewidth]{#1}}{%
      \fbox{\begin{minipage}[c][#2][c]{0.97\linewidth}\centering\small
        Missing figure: \texttt{\detokenize{figures/#1}} (.pdf or .png)
      \end{minipage}}}}%
}

\newcommand{\papertitle}{Detecting Errors in Large Language Model Responses:\\The Incremental Predictive Value of Hidden States}
\newcommand{\paperabstract}{Although large language models support a wide range of tasks, hallucinations remain a major obstacle to user trust. When a model generates an answer, both its internal processing and final output vary with the question. Internal states are often expected to offer richer signals for hallucination detection. However, studies using these states typically report their standalone discrimination performance, while output-side comparisons often use simple baselines with weaker modeling capacity. We integrate output-side information available at inference time into a combined baseline and compare detection performance before and after adding hidden states under identical data splits and evaluation rules. Experiments cover open-ended generation and constrained multiple-choice tasks across different models. Adding hidden states improves AUROC by approximately 0.004 and 0.039 on the two tasks, respectively. Hidden states thus provide additional predictive information, but strong standalone detection performance can still yield only limited additional gains.}

\title{\papertitle}
\author{Ningzhi Li \quad Xiaojun Ma \quad Xiaofei Wang \\
  Beijing Institute of Computer Technology and Application \\
  Beijing, China \\
  \texttt{inobrain0.0@gmail.com}}
\date{}
\begin{document}
\maketitle
\begin{abstract}
\paperabstract
\end{abstract}
\input{sections/01_introduction}

\input{sections/02_related_work}

\input{sections/03_method}

\input{sections/04_experimental_setup}

\input{sections/06_results}

\input{sections/08_conclusion}

\bibliography{references/references}
\clearpage
\input{sections/10_appendix}

\end{document}

%% file: sections/01_introduction.tex
\section{Introduction}
\label{sec:introduction}

Large language models (LLMs) have become widely used as interactive assistants \citep{chatterji2025}. Users obtain knowledge and information through natural-language questions, and retrieval-based systems generate answers supported by web sources \citep{nakano2021}. A central requirement is that these models answer accurately and reliably without producing false information \citep{lin2022truthfulqa,nakano2021}. Yet fluent responses can depart from facts or contradict the given context. This phenomenon is commonly called hallucination. For example, a model may fabricate a reference or add unsupported information when summarizing a document. Fluency alone therefore provides an unreliable basis for assessing factuality, motivating hallucination detection.

Prior work detects hallucinations using information available at the model's output and hidden states recorded during generation \citep{snyder2024,azaria2023}. When decoding scores are accessible, logits and token probabilities can be retained at each generation step; statistics derived from them have been used to estimate uncertainty \citep{snyder2024,vashurin2025}. Agreement among responses sampled for the same question also provides evidence of reliability \citep{manakul2023,kadavath2022}. Semantic entropy further accounts for differences in meaning to estimate uncertainty and detect some factual hallucinations \citep{kuhn2023,farquhar2024}. These findings show that output-side information available at inference time can help predict hallucinations. Hidden states, in turn, encode internal representations during answer generation and have been used to distinguish true from false statements or detect factual errors in generated answers \citep{azaria2023,snyder2024}. However, probe performance depends on classifier capacity and control settings, so high predictive performance alone does not establish the role of an internal representation \citep{hewitt2019,belinkov2022}. Richer hidden representations need not yield larger improvements once output-side information is used. Noise may impede access to useful signals, while overlap with output-side information means that some predictive capability is already available. Thus, even when hidden states distinguish correct from incorrect answers well, their standalone performance does not reveal which capabilities remain useful beyond existing output-side information. This motivates our question: after using the available output-side information, can hidden states further improve error detection, and how large is the additional gain?

We jointly model several types of output-side information as a combined baseline, check whether it preserves the predictive performance of each component block, and compare detection performance before and after adding hidden states. The difference measures their incremental predictive value relative to this baseline. Experiments on open-ended generation and constrained multiple-choice tasks examine this value under different output conditions. On open-ended generation, Qwen's standalone-statement adaptation yields a gain of 0.0043, while Llama gains 0.0040 over the combined baseline. On multiple-choice questions, Llama retains a gain of 0.0387 after jointly including the selected option text and the full response in the output baseline, with a 95\% confidence interval of $[0.0264,0.0515]$. Hidden states can therefore further improve error detection, but strong standalone performance does not necessarily yield a large additional gain. Standalone detection performance is consequently insufficient to establish their added value over output-side information.

Our contributions are:
\begin{enumerate}
  \item We formulate the question of conditional gain from hidden states, shifting attention from their standalone detection performance to their added predictive value over available non-activation information.
  \item We develop and apply an experimental and identification protocol fixed before analysis, using a combined baseline and blockwise comparisons to establish how incremental results should be interpreted.
  \item Cross-model and readout comparisons show that hidden states provide gains beyond a combined output-side baseline, while strong standalone performance need not translate into large additional improvements.\ifdefined\ANONYMOUSREVIEW\else\footnote{\mbox{\url{https://github.com/in0brain/attention_k}}}\fi
\end{enumerate}

%% file: sections/02_related_work.tex
\section{Related Work}
\label{sec:related-work}

\citet{chakraborty2025} classify hallucination detectors as black-box, gray-box, or white-box according to their access to the generating model. Black-box methods use generated text; gray-box methods also use decoding signals such as token probabilities and sequence likelihoods; white-box methods access internal representations such as hidden states and attention. Within approaches that do not rely on external knowledge bases or auxiliary evaluation models, information used by the first two categories can be grouped as output-side information. We therefore compare output-side information with hidden states, focusing primarily on factual errors.

\paragraph{Output-side detection.}
Output-side detectors assess reliability through decoding confidence and generation behavior. Token probabilities, sequence likelihoods, and generation entropy capture uncertainty and have been used for confidence estimation and hallucination detection \citep{kadavath2022,fadeeva2023,vashurin2025}. Signals from a single decoding pass provide limited evidence about stability across possible responses. SelfCheckGPT addresses this by checking factual consistency between a target response and repeatedly sampled responses \citep{manakul2023}. Since different phrasings may express the same meaning, semantic entropy groups semantically equivalent responses before measuring dispersion across meanings \citep{farquhar2024}. Beyond probabilities and consistency, \citet{kadavath2022} study models' direct assessments of answer correctness, and \citet{pedapati2025} investigate confidence estimation under black-box access. These studies primarily compare output-side measures, including response consistency and semantic uncertainty, and their detection performance \citep{manakul2023,farquhar2024,vashurin2025}. They do not directly estimate the conditional gain from hidden states over a baseline that jointly models multiple types of output information.

\paragraph{Hidden-state detection.}
SAPLMA trains a multilayer perceptron (MLP) on intermediate activations for true and false statements, demonstrating that hidden states can distinguish the two \citep{azaria2023}. \citet{chwang2024} further compare probe performance across layers and readout methods. Because supervised probes require labeled examples, later work explores unsupervised detection. MIND exploits statistical patterns in hidden states, while HaloScope learns useful internal representations from unlabeled model generations \citep{su2024,du2024}. Other work extends consistency analysis to internal representations. INSIDE measures reliability through differences among hidden representations from repeated generations \citep{chen2024}; DRIFT uses internal representational inconsistency to assess factuality \citep{bhatnagar2026}; and semantic entropy probes predict semantic uncertainty directly from hidden states, reducing the computation required for repeated sampling \citep{kossen2024}. \citet{han2025} apply lightweight probes to long-form factuality detection and compare their performance and computational cost with methods requiring repeated sampling. \citet{orgad2025} examine cross-dataset generalization and find limited additional generalization over an output-logit baseline. The former compares against repeated-sampling detectors, and the latter controls for a single output-side signal. Neither design establishes how much predictive information hidden states add after multiple output-side sources are jointly modeled.

Some studies evaluate several detection signals within a common framework. \citet{snyder2024} use a unified classifier to compare output probabilities, input attributions, self-attention, and feed-forward activations for factual error detection. LLM-Check uses an auxiliary language model's hidden states, attention, and output probabilities to detect hallucinations in a single response \citep{sriramanan2024}. Probe performance nevertheless depends on classifier capacity, training, and controls \citep{hewitt2019,voita2020,pimentel2020,belinkov2022}, and non-target dataset cues can also produce high detection performance \citep{hussain2026}. Strong performance from hidden states, alone or combined with other signals, is therefore insufficient to establish their added value. We use a combined baseline of multiple output-side sources to assess whether hidden states provide additional predictive gains.

%% file: sections/03_method.tex
\section{Method}
\label{sec:method}

\subsection{Overall Design and Conditional Gain}
\label{sec:delta-definition}

We evaluate the predictive value that hidden states add to existing output-side information. To account for the predictive capabilities of multiple output-side sources, we construct a combined baseline and use blockwise comparisons to check whether it preserves each block's performance. We then hold the data splits, learning procedure, and output-side prediction scores fixed, compare detection performance before and after adding hidden states, and check whether the fused model outperforms each block. Because additional gains also depend on output-side coverage and hidden-state readout, we vary these settings separately to examine both dependencies.

Standalone prediction from hidden states evaluates their detection capability; the performance difference between the combined baseline and the fused model evaluates their added value.

Let $Y\in\{0,1\}$ denote the error label of a generated response, and let $S$ denote the non-activation information allowed by the protocol, including prompts, generated outputs, repeated-generation statistics, and accessible decoding signals. We use \emph{output-side information} as an umbrella term that also includes prompt-side controls. Let $H$ denote hidden states extracted using a predetermined layer and pooling rule.

We measure error prediction performance by the area under the receiver operating characteristic curve (AUROC). The combined baseline $\Sstar$ is constructed from $S$, and adding the hidden-state readout yields the fused model $\Sstar+H$. We define the conditional gain as
\begin{equation}
  \deltah
  =\operatorname{AUROC}(\Sstar+H)-\operatorname{AUROC}(\Sstar).
  \label{eq:delta-h}
\end{equation}
Under the current experimental setting, $\deltah>0$ indicates that adding hidden states improves error detection over the combined output-side baseline.

\Cref{tab:method-notation} summarizes the notation.
\input{tables/table_method_notation}

Both models use this conditional comparison; \cref{fig:framework} illustrates the main procedure.
\begin{paperfigure}
  \centering
  \resizebox{\textwidth}{!}{

\input{figures/figure_01_framework}}
  \caption{Framework for evaluating conditional gain.}
  \label{fig:framework}
\end{paperfigure}

\subsection{Combined Baseline and Hidden-State Fusion}
\label{sec:baseline-fusion}

\paragraph{Information blocks.}
Different output-side sources offer different cues for error prediction. To evaluate hidden states after jointly using these cues, the baseline contains five predefined blocks covering the answer, generation behavior, and prompt context. Both tasks use the same information categories, with representations adapted to their output formats:
\begin{equation}
  \Sstar=g(B_1,B_2,B_3,B_4,B_5).
  \label{eq:stacked-observable-baseline}
\end{equation}
\Cref{tab:block-overview} specifies the blocks for each task.
\input{tables/table_block_overview}

\paragraph{Two-level stacking.}
Because the blocks have different representations, we first fit a base learner to each block and then combine their prediction scores through two-level stacking \citep{wolpert1992}. The baseline and fused model share the output-side scores; the fused model receives only one additional column of hidden-state scores. This evaluates improvement over the same output-side predictions.

The meta-learner is trained on cross-fitted scores within each outer training fold, avoiding combinations learned from base learners' in-sample fitted scores. At test time, the base learners score the shared outer test fold, and the meta-learner produces the final predictions. All preprocessing and hyperparameter selection are confined to training folds \citep{varma2006}. The \emph{Within-fold fitting} and \emph{Stacking and testing} stages of \cref{fig:framework} illustrate training and prediction.

\subsection{Blockwise Comparisons and Criteria}
\label{sec:blockwise-criteria}

Joint modeling requires checking whether individual blocks' predictive capabilities are preserved. If the combined baseline performs worse than a single block, an improvement from fusion may still leave the fused model below that block. We therefore assess baseline preservation and fused-model improvement relative to each block separately.

To check whether the baseline preserves each block's performance, we first compute its difference from the single-block predictor $B_k$:
\begin{equation}
  \Delta_{S,k}
  =\operatorname{AUROC}(\Sstar)-\operatorname{AUROC}(B_k).
  \label{eq:baseline-block-gap}
\end{equation}
The non-inferiority check allows an AUROC decrease of $\varepsilon_{NI}=0.02$ relative to each block and requires every one-sided lower confidence bound for $\Delta_{S,k}$ to exceed $-\varepsilon_{NI}$. This tolerance was fixed before the formal analysis first accessed the true labels.

To determine whether adding hidden states yields performance above each block, we also compute
\begin{equation}
  \Delta_{D,k}
  =\operatorname{AUROC}(\Sstar+H)-\operatorname{AUROC}(B_k).
\end{equation}
The direct blockwise comparison requires every one-sided lower bound for $\Delta_{D,k}$ to exceed zero. Passing all comparisons establishes that the fused model outperforms every block included in the comparison.

Non-inferiority allows a small baseline loss relative to individual blocks. To assess whether the hidden-state gain exceeds this allowed loss, we add a more conservative consistency check: the baseline must pass non-inferiority, and the lower confidence bound for $\Delta_H$ must exceed $\varepsilon_{NI}$.

%% file: tables/table_method_notation.tex
\begin{papertable}
  \centering
  \caption{Main notation.}
  \label{tab:method-notation}
  \small
  \begin{tabularx}{\textwidth}{@{}l>{\raggedright\arraybackslash}X@{}}
    \toprule
    Symbol & Meaning \\
    \midrule
    $Y$ & Error label: 1 for an erroneous response and 0 otherwise \\
    $S$ & Output-side information included in this study, including prompt-side controls \\
    $B_k$ & Information block $k$ ($k=1,\ldots,5$); its predictor in performance comparisons \\
    $H$ & Hidden representation from a prespecified layer and pooling rule; its predictor when standalone performance is reported \\
    $\Sstar$ & Combined baseline constructed from the five information blocks \\
    $\Sstar+H$ & Fused model adding the hidden-state readout to the combined baseline \\
    $\deltah$ & AUROC gain of the fused model over the combined baseline \\
    $\Delta_{S,k}$ & AUROC difference between the combined baseline and block-$k$ predictor \\
    $\Delta_{D,k}$ & AUROC difference between the fused model and block-$k$ predictor \\
    $\varepsilon_{NI}$ & Prespecified tolerance for baseline loss in the non-inferiority check \\
    \bottomrule
  \end{tabularx}
\end{papertable}

%% file: figures/figure_01_framework.tex
\begin{tikzpicture}[
  font=\sffamily\fontsize{6.6}{8.4}\selectfont,
  >={Stealth[length=1.7mm,width=1.3mm]},
  box/.style   ={draw=black!55, rounded corners=1.5pt, align=left,
                 inner xsep=4pt, inner ysep=3.5pt, minimum height=5.4mm},
  blk/.style   ={box, fill=white,   text width=25mm},
  proc/.style  ={box, fill=black!3, text width=26mm},
  base/.style  ={box, fill=black!9, text width=25mm},
  hid/.style   ={box, fill=white,   text width=25mm, draw=black!45, dashed},
  res/.style   ={box, fill=white,   text width=25mm},
  gain/.style  ={box, fill=black!9, text width=29mm},
  stage/.style ={align=left, font=\sffamily\fontsize{6.6}{8.4}\selectfont\bfseries, text=black!65},
  note/.style  ={font=\sffamily\fontsize{6.6}{8.4}\selectfont, text=black!55, align=left},
  flow/.style  ={->, draw=black!60, line width=0.5pt},
  audit/.style ={->, draw=black!45, line width=0.5pt, dashed},
]

\def\GXa{0}
\def\GXb{3.5}
\def\GXc{7.1}
\def\GXd{10.7}
\def\GXe{14.2}

\def\GYtop{3.75}
\def\GYs{1.00}
\def\GYh{-2.20}
\def\GYband{-5.30}

\node[proc] (sample) at (\GXa,\GYs+0.50)
  {\textbf{Two tasks}\\H1: open-ended recall\\MCQ: multiple choice\\Sec.~\ref{sec:task-data}};
\node[proc] (split)  at (\GXa,\GYs-1.15)
  {\textbf{Shared outer splits}\\Prespecified groups\\Sec.~\ref{sec:evaluation-protocol}};
\draw[flow] (sample) -- (split);

\node[note] (slab) at (\GXb, 3.00) {\textbf{Output-side information $S$}};
\node[blk] (b1) at (\GXb, 2.36) {$B_1$\ \ Answer identity};
\node[blk] (b2) at (\GXb, 1.68) {$B_2$\ \ Length and format};
\node[blk] (b3) at (\GXb, 1.00) {$B_3$\ \ Output text};
\node[blk] (b4) at (\GXb, 0.32) {$B_4$\ \ Scores and agreement};
\node[blk] (b5) at (\GXb,-0.36) {$B_5$\ \ Prompt context};

\begin{scope}[on background layer]
  \node[draw=black!30, rounded corners=2pt, fit=(slab)(b1)(b5),
        inner sep=3.5pt, fill=black!2] (sbox) {};
\end{scope}

\node[hid] (h) at (\GXb,\GYh)
  {\textbf{Hidden states $H$}\\Fixed layer and pooling\\Sec.~\ref{sec:model-training}};

\node[proc] (fit) at (\GXc,\GYs)
  {\textbf{Fit each block}\\Select learners and parameters in training folds};
\node[proc] (read) at (\GXc,\GYh)
  {\textbf{Hidden-state readout}\\Same splits and selection rules};

\draw[flow] (sbox.east) -- (fit.west);
\draw[flow] (h) -- (read);
\draw[flow] (split.east)  .. controls +(0.8,0) and +(-0.9,0) .. (sbox.west);
\draw[flow] (split.south) .. controls +(0,-0.9) and +(-0.9,0) .. (h.west);

\node[base] (sstar) at (\GXd,\GYs)
  {\textbf{Shared output scores}\\One column per block\\Cross-fitted for training};
\node[proc] (hm) at (\GXd,\GYh) {\textbf{Hidden-state scores}\\Additional input to fusion};

\draw[flow] (fit) -- (sstar);
\draw[flow] (read) -- (hm);

\node[res]  (aus)   at (\GXe, 1.62) {\textbf{Combined baseline $\Sstar$}\\Meta-learner: five scores\\$\operatorname{AUROC}(\Sstar)$};
\node[res]  (aush)  at (\GXe, 0.05) {\textbf{Fused model $\Sstar+H$}\\Meta-learner: six scores\\$\operatorname{AUROC}(\Sstar{+}H)$};
\node[gain] (delta) at (\GXe,-2.05)
  {\textbf{Conditional gain $\deltah$}\\$\operatorname{AUROC}(\Sstar{+}H)-\operatorname{AUROC}(\Sstar)$\\Definition: Sec.~\ref{sec:delta-definition}\\Results: Sec.~\ref{sec:main-increment}};

\draw[flow] (sstar.east) .. controls +(0.9,0) and +(-0.9,0) .. (aus.west);
\draw[flow] (sstar.east) .. controls +(0.9,0) and +(-0.9,0) .. (aush.west);

\draw[flow] (hm.north)   .. controls +(0,0.9) and +(-1.1,0) .. (aush.west);
\draw[flow] (aus.east)   .. controls +(0.6,0) and +(0.6,0)  .. (delta.east);
\draw[flow] (aush.south) -- (delta.north);

\node[box, text width=70mm] (basecheck) at (3.1,-3.85)
  {\textbf{Baseline vs.\ individual blocks: $\Delta_{S,k}$}\\Check preservation of each block's predictive performance\\Criteria: Sec.~\ref{sec:blockwise-criteria}};
\node[box, text width=70mm] (fusioncheck) at (11.1,-3.85)
  {\textbf{Fusion vs.\ individual blocks: $\Delta_{D,k}$}\\Check whether fusion outperforms each block\\Criteria: Sec.~\ref{sec:blockwise-criteria}};
\draw[audit] (aus.south west) -- (12.50,0.98) -- (12.50,-3.12) -| (basecheck.north east);
\draw[audit] (aush.east) .. controls +(1.8,0) and +(1.8,0) .. (fusioncheck.east);

\node[box, fill=white, draw=black!40, text width=158mm] (bound) at (7.1,\GYband)
  {\textbf{Factors affecting gains}\quad Output coverage; hidden-state readout and representation (Sec.~\ref{sec:supplementary-setup})\\
   Numerical handling and grouping sensitivity (Sec.~\ref{sec:evaluation-protocol})};

\node[stage, anchor=west] at (sample.west |- 0,\GYtop) {Inputs and splits};
\node[stage, anchor=west] at (sbox.west   |- 0,\GYtop) {Information sources\\(\ref{sec:baseline-fusion}, \ref{sec:model-training})};
\node[stage, anchor=west] at (fit.west    |- 0,\GYtop) {Within-fold fitting\\(\ref{sec:baseline-fusion})};
\node[stage, anchor=west] at (sstar.west  |- 0,\GYtop) {Shared prediction scores\\(\ref{sec:baseline-fusion})};
\node[stage, anchor=west] at (aus.west    |- 0,\GYtop) {Stacking and testing};
\draw[draw=black!25, line width=0.4pt]
  ([yshift=-3.2mm]sample.west |- 0,\GYtop) -- ([yshift=-3.2mm]bound.east |- 0,\GYtop);

\draw[flow] ($(aus.north west)+(0,1.10)$) -- ++(0.65,0)
  node[right=1.5mm, inner sep=0pt, text=black!65] {Prediction and evaluation};
\draw[audit] ($(aus.north west)+(0,0.60)$) -- ++(0.65,0)
  node[right=1.5mm, inner sep=0pt, text=black!65] {Blockwise checks};

\end{tikzpicture}

%% file: tables/table_block_overview.tex
\begin{papertable}
  \centering\small
  \caption{Five information blocks in the main-analysis baseline.}
  \label{tab:block-overview}
  \begin{tabularx}{\textwidth}{@{}lp{0.20\textwidth}>{\raggedright\arraybackslash}X>{\raggedright\arraybackslash}X@{}}
    \toprule
    Block & Category & H1 & MCQ \\
    \midrule
    $B_1$ & Answer identity & Eligible generated identifier strings & Selected answer letter \\
    $B_2$ & Surface form & Length, format, identifier count and positions & Option lengths and lexical overlap \\
    $B_3$ & Text content & Full generated text & Selected option text \\
    $B_4$ & Generation and sampling signals & Token scores, cross-sample consistency, and explicit confidence & Candidate-answer scores and cross-sample statistics \\
    $B_5$ & Prompt context & Prompt text, task family, and generation route & Question stem and all options \\
    \bottomrule
  \end{tabularx}
\end{papertable}

%% file: sections/04_experimental_setup.tex
\section{Experimental Setup}
\label{sec:experimental-setup}

\subsection{Tasks and Data}
\label{sec:task-data}

We examine open-ended recall (H1) and constrained multiple-choice questions (MCQ) to assess the added value of hidden states under different task settings. H1 generates text containing identifiers; MCQ asks the model to select a given option and return its answer letter.

Both tasks require explicit correctness criteria. Cybersecurity provides standardized identifier systems and question-answering datasets with reference answers. We therefore construct H1 from MITRE ATT\&CK and CWE entries \citep{strom2020,mitrecwe} and use CyberMetric for MCQ \citep{tihanyi2024}.

\paragraph{Open-ended recall (H1).}
The model generates an identifier from the name and description of a technique or weakness. The task detects fabricated identifiers by checking their existence in a fixed knowledge base. After excluding identifiers copied verbatim from the prompt, a response is labeled erroneous if any remaining ATT\&CK or CWE identifier is absent from the knowledge base. Each generation record is one analysis unit.

\paragraph{Constrained multiple choice (MCQ).}
The model selects an answer from the given options. A selection that differs from the reference answer is labeled erroneous. Each question is one analysis unit.

For both models, only samples with valid labels and hidden states are analyzed; the two analysis pipelines are called the H1 and MCQ arms. Qwen includes 8,664 H1 and 1,750 MCQ records; Llama includes 8,810 and 1,746, respectively\appendixnote{note:appendix-samples}{See Appendix~\ref{app:setup}.}. H1 generations from the same prompt remain in the same group, while each MCQ question forms its own group. Training and test sets are split by group, as shown in the \emph{Inputs and splits} stage of \cref{fig:framework}.

\subsection{Models and Training}
\label{sec:model-training}

We use Qwen2.5-7B-Instruct \citep{qwen2024} and Llama-3.1-8B-Instruct, both with a local 8-bit quantized backend. Each model uses the same learning procedure and sampling parameters across tasks. To apply a common layer-selection rule across architectures, we select the hidden-state layer at a fixed fraction of total model depth, set in advance to \(0.7\). H1 averages hidden vectors over all eligible identifier tokens; MCQ uses the vector at the answer-letter position.

Text blocks use word- and character-level term frequency--inverse document frequency (TF-IDF) features; dense blocks and hidden states are standardized. Base learners and meta-learners use L2-regularized logistic regression. Base-learner regularization is selected by grouped cross-validation within each outer training fold. Feature extraction and predictor training correspond to \emph{Information sources} and \emph{Within-fold fitting} in \cref{fig:framework}.

\subsection{Output Coverage and Readout Comparisons}
\label{sec:supplementary-setup}

The output-coverage comparison tests whether hidden states remain useful when more response text is incorporated. Within each model, we hold the analyzable samples, splits, hidden states, and statistical protocol fixed and vary only the MCQ text block $B_3$. The three conditions use the selected option text, the full model response, and their concatenation. Text representations and predictors are refitted within training folds for each condition.

Readout comparisons test whether different learners can make fuller use of hidden-state signals. Using Qwen's existing vectors and output-side scores, we compare L2 tuning ranges, MLPs, L1 logistic regression, and XGBoost. H1 representation adaptation converts each distinct identifier into a standalone statement about its existence in the catalog, extracts new statement representations, trains on statement labels, and aggregates predictions to generation records using a prespecified maximum score.

\input{sections/05_evaluation_protocol}

%% file: sections/05_evaluation_protocol.tex
\subsection{Evaluation and Sensitivity Comparisons}
\label{sec:evaluation-protocol}

\paragraph{Primary evaluation.}
The combined baseline and fused model are evaluated on the same outer test folds to compare performance on identical samples. Both models use the grouping described above, with five outer folds, three inner folds, and three repetitions. We compute the out-of-fold AUROC difference within each repetition and then average the differences.

We estimate 95\% confidence intervals using a paired grouped bootstrap \citep{field2007}, comparing predictors on the same resampled records. Each MCQ question contributes one evaluated response, so sampling a question includes that response. Each H1 prompt has multiple responses; sampling a prompt group includes all its responses to preserve their within-group dependence. A question or prompt group may be sampled more than once.

\paragraph{Information and numerical controls.}
To examine dependence on generation and sampling signals, both models compare path \pathA, which retains all five blocks, with path \pathC, which removes $B_4$ from both the baseline and fused model. Qwen's H1 $B_4$ has a non-finite-value handling issue. Path \pathB therefore reconstructs the affected features while holding hidden states fixed to examine the effect of numerical handling.

\paragraph{Grouping sensitivity.}
The main H1 split allows identifiers to be shared between training and test sets. To examine dependence on such sharing, Qwen is also evaluated using groups defined by identifiers, vocabulary blocks, and joint prompt--fabricated-identifier relationships. The vocabulary-block comparison redefines labels and hidden states by block; the other comparisons retain the original prediction task.

%% file: sections/06_results.tex
\section{Results}
\label{sec:results}

\subsection{Additional Predictive Gains from Hidden States}
\label{sec:main-increment}
\label{sec:llama-results}

To assess whether hidden states improve error detection beyond output-side information, we compare the combined baseline, standalone hidden-state predictor, and fused model for Qwen and Llama. \Cref{tab:e1-results} reports these results alongside readout and representation comparisons.
\input{tables/table_02_e1}

Both models improve on both tasks. On H1, AUROC increases by $0.0016$ for Qwen and $0.0040$ for Llama; on MCQ, the gains are $0.0289$ and $0.0596$, respectively\appendixnote{note:appendix-results}{See Appendix~\ref{app:results}.}. All four main-result confidence intervals lie above zero, showing that hidden states provide additional predictive information.

Both models exhibit the same pattern: standalone hidden-state detection is stronger on H1 than on MCQ, but its improvement over the output baseline is smaller. H1 labels depend on identifier existence in a fixed knowledge base. Using identifier identity alone, Qwen already achieves an AUROC of $0.9826$\appendixnoteref{note:appendix-results}, indicating strong predictive cues from identity. The combined baseline approaches the AUROC ceiling, leaving little room for absolute improvement. Strong standalone detection can therefore yield only limited additional gains.

Both models pass baseline non-inferiority on both tasks, and their fused models outperform every compared block. The more conservative consistency check passes only for Llama's main MCQ result: its gain interval has a lower bound of $0.0447$, above $0.02$. The other three settings do not reach this threshold\appendixnoteref{note:appendix-results}.

\subsection{Effect of Output-Side Information}
\label{sec:output-information-results}

To explain how conditional gain changes with the output baseline, we first compare retaining and removing generation and sampling signals. For both models, removing these signals increases the conditional gain while lowering fused-model performance. \Cref{fig:b4-dependency} shows the Qwen comparison: removing $B_4$ reduces baseline performance more than fused-model performance on both tasks, widening their gap. The larger gain thus reflects a weaker output baseline, while final detection performance decreases.

\begin{paperfigure}
  \centering
  \includegraphics[width=\linewidth]{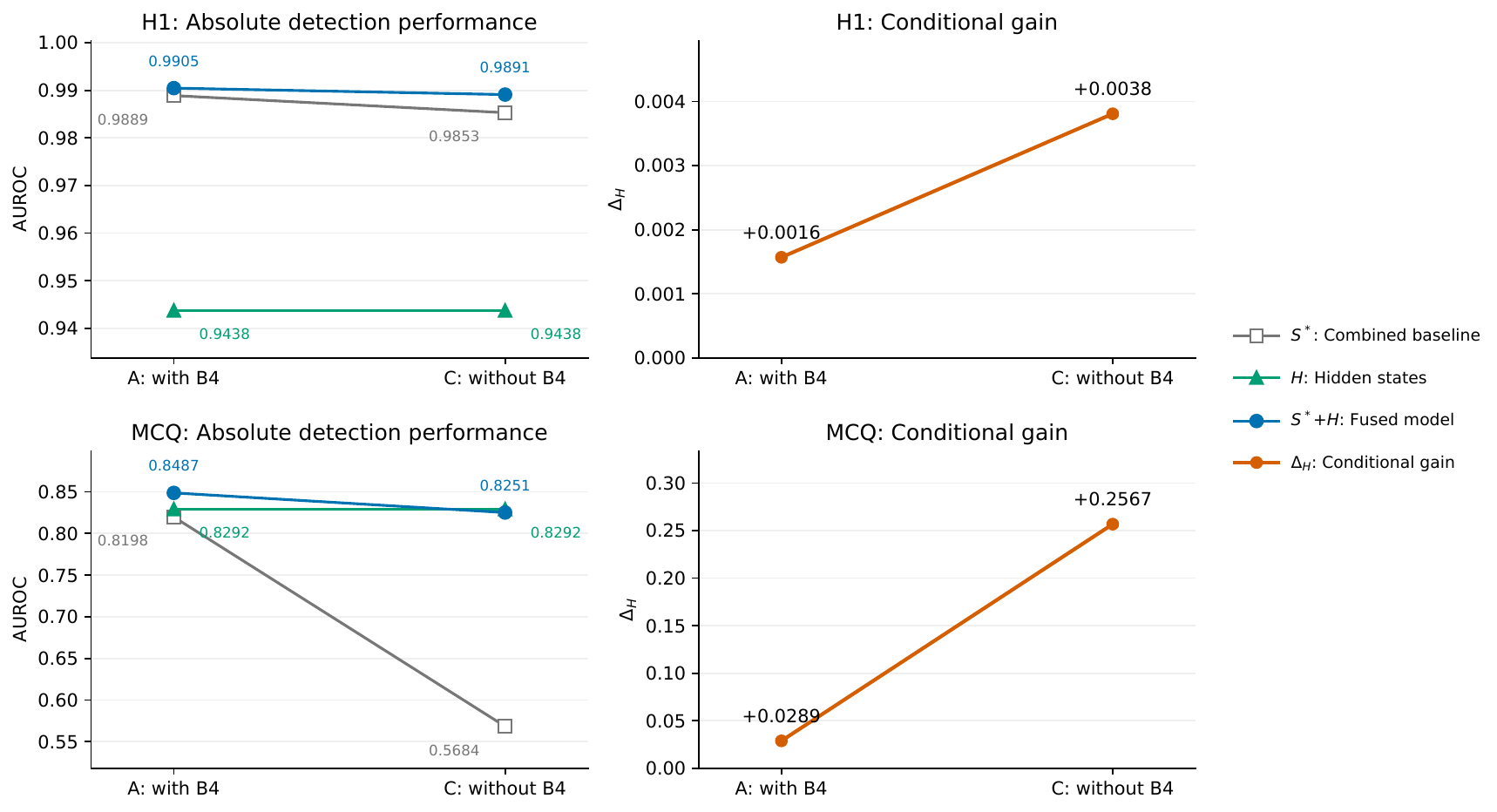}
  \caption{Absolute performance and conditional gain with and without $B_4$. Each panel has its own vertical scale; points and connecting lines show point estimates only.}
  \label{fig:b4-dependency}
\end{paperfigure}

In the numerical control, reconstructing the affected features yields Qwen gains of $0.0028$ on H1 and $0.0273$ on MCQ, preserving the positive direction and respective magnitudes.

We next jointly include the selected option text and the full response in the MCQ baseline (\cref{tab:mcq-output-coverage}). Qwen typically returns an answer letter, whereas Llama produces explanatory text; joint encoding therefore adds more previously uncovered text for Llama. Its baseline AUROC rises from $0.7465$ to $0.7689$, while fused performance changes from $0.8061$ to $0.8075$. The larger baseline improvement reduces the hidden-state gain from $0.0596$ to $0.0387$. Qwen's gain changes only slightly, from $0.0289$ to $0.0277$.
\input{tables/table_mcq_output_coverage}

Both models retain positive gains after the full response is added, while improvements to the output baseline reduce the additional benefit from hidden states. The gain thus depends both on the improvement contributed by hidden states and on how fully existing output-side cues are used.

\subsection{Effect of Readout and Representation}
\label{sec:probe-sensitivity}

We examine the readout and representation comparisons in \cref{tab:e1-results} to assess whether better access to hidden-state signals increases their added value. With alternative readouts and wider regularization searches, H1 gain estimates range from $-0.0013$ to $0.0024$. Standalone-statement adaptation raises hidden-state AUROC from $0.9438$ to $0.9780$, but increases the gain over the output baseline only from $0.0016$ to $0.0043$. The improvement in standalone detection does not translate into an equally large additional gain, as the strong output baseline continues to limit absolute headroom.

MCQ gains are more sensitive to readout choice. L2, L1, and XGBoost yield positive improvements, whereas the MLP gain is $0.0045$ with a confidence interval containing zero. These differences occur under the same output baseline, indicating that the readout affects how effectively the additional predictive capability of hidden states is used.

%% file: tables/table_02_e1.tex

\begin{papertable}
  \centering\linespread{1}\small
  \setlength{\tabcolsep}{4pt}
  \renewcommand{\arraystretch}{1.0}
  \setlength{\aboverulesep}{0.3ex}
  \setlength{\belowrulesep}{0.4ex}
  \caption{Error detection for both models, with readout and representation comparisons.}
  \label{tab:e1-results}
  \label{tab:probe-sensitivity}
  \label{tab:probe-full}
  \begin{tabular*}{0.98\linewidth}{@{\extracolsep{\fill}}clrrrl@{}}
    \toprule
    Model & \multicolumn{1}{c}{Task} & $\Sstar$ & $H$ & $\Sstar+H$ & $\deltah$ (95\% CI) \\
    \midrule
    \multirow{2}{*}{Qwen} & \multicolumn{1}{c}{H1} & 0.9889 & 0.9438 & 0.9905 & $+0.0016$ $[+0.0006,+0.0028]$ \\
     & \multicolumn{1}{c}{MCQ} & 0.8198 & 0.8292 & 0.8487 & $+0.0289$ $[+0.0189,+0.0389]$ \\
    \cmidrule(lr){1-6}
    \multirow{2}{*}{Llama} & \multicolumn{1}{c}{H1} & 0.9712 & 0.9148 & 0.9752 & $+0.0040$ $[+0.0014,+0.0075]$ \\
     & \multicolumn{1}{c}{MCQ} & 0.7465 & 0.7916 & 0.8061 & $+0.0596$ $[+0.0447,+0.0751]$ \\
    \midrule
    Task & Readout or representation & $\Sstar$ & $H$ & $\Sstar+H$ & $\deltah$ (95\% CI) \\
    \midrule
    \multirow{6}{*}{H1} & L2, wide grid & \multirow{6}{*}{0.9889} & 0.9482 & 0.9910 & $+0.0021$ $[+0.0008,+0.0037]$ \\
     & MLP adaptation &  & 0.9330 & 0.9913 & $+0.0024$ $[+0.0009,+0.0043]$ \\
     & L1 adaptation &  & 0.9328 & 0.9906 & $+0.0017$ $[+0.0006,+0.0029]$ \\
     & XGBoost adaptation &  & 0.9586 & 0.9910 & $+0.0021$ $[+0.0009,+0.0035]$ \\
     & L1, wide grid &  & 0.9193 & 0.9876 & $-0.0013$ $[-0.0039,+0.0011]$ \\
     & Standalone statements &  & 0.9780 & 0.9931 & $+0.0043$ $[+0.0020,+0.0070]$ \\
    \cmidrule(lr){1-6}
    \multirow{5}{*}{MCQ} & L2, wide grid & \multirow{5}{*}{0.8198} & 0.8292 & 0.8487 & $+0.0289$ $[+0.0189,+0.0389]$ \\
     & MLP adaptation &  & 0.7826 & 0.8242 & $+0.0045$ $[-0.0005,+0.0095]$ \\
     & L1 adaptation &  & 0.8030 & 0.8445 & $+0.0247$ $[+0.0151,+0.0345]$ \\
     & XGBoost adaptation &  & 0.8085 & 0.8426 & $+0.0228$ $[+0.0134,+0.0326]$ \\
     & L1, wide grid &  & 0.8030 & 0.8445 & $+0.0247$ $[+0.0151,+0.0345]$ \\
    \bottomrule
  \end{tabular*}

  \vspace{0.35em}
  \begin{minipage}{0.98\linewidth}
    \footnotesize
    Model comparisons use the original L2 grid; readout and representation comparisons use Qwen. The baseline contains five output-side blocks. The first three numerical columns report AUROC averaged over three repetitions. Readout comparisons share output-side scores; statement adaptation aggregates statement predictions to generation records before evaluation.
  \end{minipage}
\end{papertable}

%% file: tables/table_mcq_output_coverage.tex
\begin{papertable}
  \centering\small
  \setlength{\tabcolsep}{4pt}
  \caption{Output coverage of the MCQ text block.}
  \label{tab:mcq-output-coverage}
  \begin{tabular}{llrrrl}
    \toprule
    Model & $B_3$ source & $\Sstar$ & $H$ & $\Sstar+H$ & $\deltah$ (95\% CI) \\
    \midrule
    Qwen & Option text & 0.8198 & 0.8292 & 0.8487 & $+0.0289$ $[+0.0189,+0.0389]$ \\
    Qwen & Full response & 0.8221 & 0.8292 & 0.8495 & $+0.0274$ $[+0.0172,+0.0378]$ \\
    Qwen & Option + response & 0.8221 & 0.8292 & 0.8498 & $+0.0277$ $[+0.0176,+0.0379]$ \\
    \midrule
    Llama & Option text & 0.7465 & 0.7916 & 0.8061 & $+0.0596$ $[+0.0447,+0.0751]$ \\
    Llama & Full response & 0.7699 & 0.7916 & 0.8076 & $+0.0377$ $[+0.0255,+0.0502]$ \\
    Llama & Option + response & 0.7689 & 0.7916 & 0.8075 & $+0.0387$ $[+0.0264,+0.0515]$ \\
    \bottomrule
  \end{tabular}
  \vspace{0.3em}
  \begin{minipage}{0.97\linewidth}
    \footnotesize Samples, splits, and $H$ are fixed within each model. The first three numerical columns report AUROC; intervals are 95\% confidence intervals.
  \end{minipage}
\end{papertable}

%% file: sections/08_conclusion.tex
\section{Conclusion and Discussion}
\label{sec:conclusion}
\label{sec:discussion}

We evaluate the added predictive value of hidden states using a combined baseline of multiple output-side sources and blockwise comparisons. Hidden states provide predictive information beyond the available output-side information, but strong standalone detection does not necessarily yield large additional gains. On H1, the combined baseline already performs well, improvements from hidden states are small, and the tested readout and representation adaptations preserve this pattern. On MCQ, hidden states continue to add predictive value over the combined output-side baseline. Standalone detection performance is therefore insufficient to establish their added value over existing output-side information.

Baseline construction and information coverage affect gain estimates. Removing generation and sampling signals can increase the gain while reducing fused-model performance. Evaluations should report baseline performance, standalone hidden-state performance, fused performance, and conditional gain together, and specify the output information being compared. These conclusions concern predictive gains under this study's tasks, models, backends, and learning procedures.

\section*{Limitations}

This study has limitations in numerical integrity, parsing conventions, and probe training coverage. The current data do not support an estimate of generalization to unseen identifiers, and the fused model's advantage over individual blocks varies with grouping\appendixnote{note:appendix-grouping}{See Appendix~\ref{app:results}.}. AUROC measures ranking performance for error prediction; practical utility also depends on detection thresholds, probability calibration, and readout cost. Future work can assess whether predictive gains translate into practical benefits under explicitly specified conditions.

%% file: sections/10_appendix.tex

\appendix

\section{Methodological Details}
\label{app:spec}

\subsection{Five Non-Activation Blocks in Each Arm}
\label{app:blocks}

The table below specifies the fields and dimensions of the five output-side blocks for H1 and MCQ.

\begin{papertable}
  \centering\small
  \caption{Implementation of $B_1$--$B_5$ in H1 and MCQ.}
  \label{tab:app-block-map}
  \begin{tabularx}{\textwidth}{c>{\raggedright\arraybackslash}X>{\raggedright\arraybackslash}X}
    \toprule
    Block & H1 & MCQ \\
    \midrule
    $B_1$ & Text representation of eligible identifier strings in the response & Four-dimensional one-hot encoding of the selected answer letter A--D \\
    $B_2$ & 9 surface and format features: length, line count, list structure, identifier count and positions & 6 surface features: selected-option length, relative option lengths, and lexical overlap \\
    $B_3$ & Full generated text & Text of the selected option \\
    $B_4$ & 14 features: identifier-token and response log-probabilities, entropy, ranks, length-normalized scores, cross-sample consistency, and explicit confidence & 6 features: margins and entropy over four candidate letters, full-vocabulary entropy, answer-token log-probability, and cross-sample consistency \\
    $B_5$ & Prompt text with prespecified task-family and generation-route indicators & Question stem and all options, excluding the model's selection \\
    \bottomrule
  \end{tabularx}
\end{papertable}

$B_1$--$B_5$ form the combined baseline $\Sstar$; adding the hidden-state block yields $\Sstar+H$. Path \pathC removes $B_4$ from both arms, so its baseline uses the remaining four blocks and its fused model adds $H$ to those blocks.

\subsection{Within-Fold Preprocessing and Nested Stacking}
\label{app:stacking}

The baseline and fused model share within-fold preprocessing and base-learner scores. We detail the feature transformations, cross-fitting procedure, and learner settings below.

Text blocks use word 1--2-gram and word-boundary character 3--5-gram TF-IDF features. Both vocabularies use \code{min\_df}=2, at most 50,000 features each, and sublinear term frequency. The two representations are concatenated and each row is L2-normalized. Vocabularies, inverse document frequencies, and all transformations are fitted only on the current training fold and then applied to the corresponding outer test fold.

Dense blocks are standardized using dimension-wise means and standard deviations from the training fold, replacing zero standard deviations with 1. The standardized matrix is divided by $\sqrt d$, where $d$ is the block dimension, to give dense blocks a scale comparable to row-normalized text features. Test folds do not contribute to means, standard deviations, vocabularies, or hyperparameter estimates. Hidden states use the same dense preprocessing.

$\Sstar$ stacks five blocks, and $\Sstar+H$ stacks six. Within each outer training fold, a base learner is fitted independently to each block. Cross-fitted training scores are used to train the meta-learner. The base learners are then refitted on the complete outer training fold and score the outer test fold; the meta-learner produces final predictions from these scores. Both models share the same base-learner cache. Within a repetition, they differ only in whether the meta-learner receives the column of $H$ scores. \Cref{tab:app-training-spec} lists fixed learning and resampling settings.

\begin{papertable}
  \centering\small
  \caption{Prespecified learning and resampling parameters.}
  \label{tab:app-training-spec}
  \begin{tabularx}{\textwidth}{l>{\raggedright\arraybackslash}X}
    \toprule
    Item & Prespecified setting \\
    \midrule
    Outer and inner splits & 5 outer folds and 3 inner folds; grouping depends on E1, E2, E3, or M2c \\
    Base learner & L2 logistic regression, \code{liblinear}, maximum 2,000 iterations \\
    Base-learner tuning & $C\in\{0.01,0.1,1,10\}$, selected by grouped inner cross-validation within outer training folds \\
    Meta-learner & L2 logistic regression, $C=1$, \code{lbfgs}, maximum 5,000 iterations, tolerance $10^{-6}$, intercept, no class weights \\
    Meta-learner inputs & Base learners' \code{decision\_function} scores, without further standardization \\
    Repetitions and seeds & 3 repetitions; outer-fold seed $20260804+100r$ \\
    Bootstrap & 20 batches of 1,000 samples (20,000 total); batch seed $20260804+7j$ \\
    \bottomrule
  \end{tabularx}
\end{papertable}

\subsection{M1b, iv-a, iv-b, and Task-Construct Checks}
\label{app:criteria}

The protocol denotes baseline non-inferiority, direct blockwise comparison, and the consistency check as M1b, iv-a, and iv-b, respectively. It also includes a task-construct check for identity and format cues.

For each block $B_k$, define
\begin{align}
  \Delta_{S,k}&=\operatorname{AUROC}(\Sstar)-\operatorname{AUROC}(B_k),\\
  \Delta_{D,k}&=\operatorname{AUROC}(\Sstar+H)-\operatorname{AUROC}(B_k).
\end{align}
The non-inferiority tolerance is fixed at $\varepsilon_{NI}=0.02$. This operational threshold specifies the allowed baseline loss and has not been externally validated. M1b checks whether each marginal one-sided 95\% lower bound for $\Delta_{S,k}$ exceeds $-\varepsilon_{NI}$ and combines the checks through an intersection--union test \citep{berger1996}. The primary procedure uses no multiplicity correction. Simultaneous lower bounds based on the maximum statistic are reported descriptively as secondary results; their nominal coverage relies on asymptotic approximation. M1b is determined by the intersection--union rule applied to the marginal checks.

iv-a checks whether every marginal one-sided lower bound for $\Delta_{D,k}$ exceeds zero, assessing whether the fused model outperforms every compared block. The more conservative iv-b check requires M1b to pass and the lower confidence bound for $\deltah$ to exceed $\varepsilon_{NI}$. The two checks are reported separately under their respective conditions.

Within a prespecified branch, the fused model is judged superior to all compared blocks only if every direct blockwise comparison passes. This is an intersection--union test \citep{berger1996}: if any component null hypothesis holds, the probability of incorrectly passing all comparisons is no greater than that component test's significance level. Thus, valid component tests require no further adjustment for the joint decision within a branch. Reported bounds are marginal bounds for individual blockwise comparisons. The six branches are reported separately without cross-branch multiplicity correction, retaining the prespecified primary path.

The task-construct check compares the full output-text block $B_3$ with identity and format blocks $B_1$ and $B_2$:
\begin{equation}
  \begin{aligned}
    \Delta_{\mathrm{task},k}
      &=\operatorname{AUROC}(B_3)-\operatorname{AUROC}(B_k),\\
    &\qquad k\in\{1,2\}.
  \end{aligned}
\end{equation}
The primary analysis uses marginal one-sided upper bounds for conservative screening. An upper bound below $0.02$ indicates that the full-text improvement over identity or format cues does not reach the prespecified threshold. Results from triggered branches describe predictive performance under the current label definition.

\section{Data and Experimental Details}
\label{app:setup}
\label{app:posthoc}

\subsection{Model, Hidden States, and Eligibility}
\label{app:hidden}

We specify Qwen's model, hidden-state extraction, sample eligibility, and final analysis populations below.

Both arms of the Qwen confirmatory analysis use Qwen2.5-7B-Instruct with a local 8-bit quantized backend. The model has 28 decoder blocks and a hidden dimension of 3,584. Relative depth $0.7$ selects the output of block 19 (zero-indexed), corresponding to \code{hidden\_states[20]} in the model interface. Model, tokenizer, extraction layer, temperature, nucleus-sampling settings, and random seeds are identical across arms. Maximum generation lengths were set by task to 384 tokens for H1 and 16 for MCQ.

H1 enters hidden-state analysis only when a response contains at least one ATT\&CK or CWE identifier not copied from the prompt. The vector $H$ averages target-layer states over all eligible identifier tokens, counting each token position once. MCQ is eligible only when the answer can be normalized to an option in A--D; $H$ is the state of the single answer-letter token. Both arms store float32 vectors of shape $(3584,)$. H1 has 9,264 final records, of which 8,664 are eligible; MCQ has 1,751 final records, of which 1,750 are eligible. \Cref{tab:population} provides the full accounting.

The 8,664 analyzable H1 records come from 1,508 prompt groups, 345 of which contain at least one error. Another 600 records remain in the final population but are ineligible: 477 only copy prompt identifiers, 122 contain no identifier, and 1 contains only a CVE identifier. These lack the primary identifier label and corresponding hidden states. The nominal MCQ population contains 1,760 questions: 9 are excluded as whole groups after runtime technical failures and 1 is lost during normalization or parsing, leaving 1,750 for analysis. This arm generated 10,520 raw units, 10,506 belonging to retained questions and 14 failed attempts kept only for auditing. Excluded questions are not replaced, and missing hidden states are not imputed. The analyzable populations contain 8,664 and 1,750 hidden-state vectors, respectively.

\input{tables/table_01_population}

\subsection{Llama Comparability and Sample Accounting}
\label{app:llama-comparability}

The cross-model Llama comparison is non-confirmatory and uses the same type of local 8-bit backend to estimate paths A and C on both tasks. Its MCQ generation budget is 512 tokens, compared with 16 for Qwen. Llama uses a corrected parser that remained unchanged during formal production. Gains are estimated separately under each model's generation budget, parser version, and analyzable population.

Both models share the same finalized H1 list of 9,264 generation records. Qwen's analyzable population has 8,664 records, 693 positives, and 1,508 groups; Llama's has 8,810 records, 339 positives, and 1,516 groups. The intersection contains 8,450 records. Qwen excludes 477 prompt-copying responses, 122 with no identifier, and 1 containing only CVE; Llama excludes 376 prompt-copying responses and 78 with no identifier.

Qwen's finalized MCQ list contains 1,751 questions after 9 runtime technical exclusions. Excluding 1 failed greedy-response parse leaves 1,750 analyzable questions and 318 positives. Llama finalizes all 1,760 questions without these technical exclusions; removing 14 failed greedy-response parses leaves 1,746 questions and 328 positives. The analyzable intersection contains 1,737 questions. These intersections are used only to audit comparability; no models were retrained on common-sample subsets.

Llama did not run tail-reconstruction path B or vocabulary-block analysis E3, and its $B_4$ numerical integrity remains unaudited. The initial A/C run retained summaries without complete per-sample predictions. The joint full-response MCQ condition was later rerun in a separate directory, retaining eight columns of out-of-fold predictions for three repetitions, fold mappings, and per-fold audits. Its summary matched the initial run.

\subsection{MCQ Output Coverage and Parsing Denominators}
\label{app:mcq-coverage}

The MCQ output-coverage comparison is non-confirmatory. The original $B_3$ uses selected option text and does not directly include the full model response. Median analyzable response length is 1 character for Qwen and 886 for Llama, while the original $B_3$ has a median length of approximately 32 characters for both. The comparison holds samples and splits fixed within each model, changes only the text source, and refits the text block and fused model. Llama's joint-condition interval was reproduced from saved predictions under the original resampling plan. This check uses the same statistical library and verifies the computation from saved predictions to metrics and intervals. Intervals for differences between conditions were not computed.

Sampling-consistency features retain successfully parsed labels from the six responses per question and use the number of successful parses as the denominator. Among Llama's 10,560 raw responses, 168 fail parsing: 14 greedy responses and 154 sampled responses. The numbers of questions with 6, 5, 4, and 3 successful parses are 1,618, 120, 18, and 4, respectively. After excluding greedy failures, 128 of the 1,746 analysis records use a denominator below 6. None of Qwen's 1,750 analysis records has this issue.

Both consistency features can be recomputed record by record from the raw data under the current rules and match their stored values. However, the original protocol fixed six responses per question without specifying how parsing failures affect the denominator. Using successful parses is a deterministic implementation choice. The effect of alternative denominators on predictive performance and hidden-state gain has not been measured.

\subsection{Readouts, Training Budgets, and Scoring}
\label{app:posthoc-probes}

The readout comparisons and standalone-statement adaptation are non-confirmatory. The first five readout conditions were fixed before that run; the wide-grid L1 condition was added after the first run, and statement adaptation was conducted separately.

To examine dependence on hidden-state readout, these analyses reuse Qwen's existing vectors, samples, and outer splits. The original L2 grid is $C\in\{0.01,0.1,1,10\}$; the wide grid has eight logarithmically spaced values from $10^{-3}$ to $10^4$. The original L1 candidates are $\{0.01,0.1,0.5,1,10\}$, with $100,1000,10000$ added after the initial run. XGBoost combines depths $\{3,6\}$ with $\{200,1000\}$ trees at learning rate $0.1$. All candidate selection is confined to training folds.

Following the SAPLMA architecture, the MLP has three hidden layers of widths $256,128,64$, with ReLU and Adam \citep{azaria2023}. Candidate epoch limits are $\{5,50\}$ and regularization coefficients are $\{10^{-5},10^{-3},10^{-1}\}$. L1 and XGBoost reuse classifier construction from the authors' implementation of \citet{han2025}, adapted to this study's single-layer input, standardization, and grouped evaluation.

Across the six readout conditions, logistic regression outputs decision scores directly. MLP and XGBoost probabilities for the positive class are clipped to $[10^{-6},1-10^{-6}]$ and converted to log-odds before fusion. This may create ties among extreme scores; the fusion layer receives the clipped and transformed scores. The full combination of raw activations, complete author-specified settings, and native scoring was not evaluated.

For H1, the original L2 grid selects its upper bound in all 60 selections; the wider grid selects $C=100$ in 54. Widening L2 changes neither the selected parameters nor predictions on MCQ. Wide-grid L1 selects larger $C$ values at a few H1 stages but reduces the overall gain estimate.

For final outer-fold MLP fits, H1 selects the 50-epoch limit in 15/15 cases; MCQ selects 5 epochs in 12/15 and 50 in 3/15. These are selected iteration limits; complete final-fit iteration logs were not retained. Among 180 validation records for 50-epoch candidates, 165 on H1 and 67 on MCQ contain at least one inner fit that does not satisfy the convergence criterion. Counts are at the candidate-validation-record level. MLP conclusions apply to this training budget and scoring convention.

\subsection{Standalone-Statement Adaptation for H1}
\label{app:statement-adaptation}

This adaptation follows the use of final-token representations of standalone statements. Each distinct primary identifier is converted into a fixed-template statement about its existence in the catalog, with an existence label from the fixed catalog. There are 10,168 statement records, 870 distinct texts, and 8,664 original H1 evaluation records. Encoding deduplicates identical texts, but training keeps each statement record as a separate row without combining weights.

Outer splits are first materialized on the original generation records; statements inherit their parent record's fold and prompt group. On the training side, statements are cross-fitted within inner folds, and the prespecified maximum raw score aggregates them back to generation records for fusion training. Test predictions use the same aggregation. The maximum raw score serves as a record-level feature for the fusion layer. Hyperparameters are selected using AUROC after record-level aggregation. This condition changes representation, training unit, and aggregation together.

Across the three repetitions, approximately $93.4\%$--$93.8\%$ of test statement records have identical text in the corresponding training fold. Results therefore describe the current prompt-grouped population, which allows repeated identifiers. All 60/60 selections reach the original grid's upper bound; other tuning ranges were not evaluated. Statement-level raw scores, fusion-training columns, and some final-fit logs were not retained.

All saved target-layer vectors and upstream checks are finite, but all 870/870 distinct texts produce non-finite values in the returned final-layer states and output scores. Hooks directly locate the anomaly within the final decoder block for two inputs; equivalent internal localization was not completed for the remaining texts.

\subsection{Repetition-Averaged Estimand and Uncertainty}
\label{app:uncertainty}

Performance differences are averaged over three repetitions, with intervals estimated by paired grouped bootstrap. Numerical error due to finite resampling is estimated separately.

Each repetition produces out-of-fold scores for $\Sstar$, $H$, and $\Sstar+H$, with AUROC first calculated within that repetition. The primary estimand is the mean AUROC difference across three repetitions:
\begin{equation}
  \begin{aligned}
    \deltah&=\frac{1}{3}\sum_{r=1}^{3}\Delta_{H,r},\\
    \Delta_{H,r}&=\operatorname{AUROC}_{r}(\Sstar+H)
      -\operatorname{AUROC}_{r}(\Sstar).
  \end{aligned}
\end{equation}

Intervals use a grouped bootstrap. The same group sample is shared across all three repetitions to preserve pairing. H1 resamples 1,508 prompt groups; MCQ resamples 1,750 questions. Every E1 branch has 20,000 valid bootstrap samples, with none discarded for containing only one class. Two-sided 95\% intervals use the 2.5th and 97.5th percentiles of the pooled bootstrap distribution. Marginal one-sided lower and upper bounds use the 5th and 95th percentiles.

Each bootstrap sample draws the original number of groups with replacement and concatenates all records in the selected groups, retaining repetitions of groups. H1's 693 error records are sampled with their groups, preserving within-group label composition. The same resampled records are used for all predictors and repetitions to estimate paired AUROC differences. Based on fixed out-of-fold predictions, the intervals quantify variation induced by group resampling. Their endpoints are the corresponding distribution percentiles, and the group count denotes the number of sampling units. Interpretation assumes independence between groups.

The 20,000 samples are split into 20 batches. Monte Carlo error in interval endpoints and widths is estimated primarily by a delete-one-batch jackknife and checked secondarily by the batch-means method. Width error bands are centered on the interval width from the pooled distribution and use $t_{0.975,19}=2.093$. These bands quantify numerical error from finite resampling. A 95\% interval width above 0.04 for $\deltah$ is classified as insufficient precision.

\subsection{Numerical Paths, Analysis Freeze, and Unblinding}
\label{app:freeze}

The numerical control holds hidden states fixed and reconstructs affected generation features. We document the checks, path settings, and data and rules fixed before formal analysis below.

Path \pathB yields an MCQ gain of $0.0273$ ($[0.0175,0.0372]$), close to path \pathA, and an H1 gain of $0.0028$ ($[0.0014,0.0046]$), still on the order of a few thousandths. This numerical check preserves the direction of E1 gains and their respective magnitudes.

The original H1 $B_4$ computation filters non-finite token log-probabilities, making features depend on a record-specific subset of valid tokens; this issue affects almost all records. Verification of path \pathB confirms that H1 E3 block-level hidden states match the original cache elementwise and that reconstructed fields are finite in both arms. Numerical handling preserves the main E1 direction while changing some E3 criteria. Path \pathB measures sensitivity to numerical handling in the original implementation.

Samples, features, splits, and interpretation rules were fixed before the formal analysis accessed true labels \citep{nosek2018}; analyses modified after unblinding are identified separately. We report protocol hashes and the unblinding time below. Appendix~\ref{app:other-estimands} describes the cross-arm contrast and confidence-stratified analysis that did not yield valid estimates.

Path \pathA follows the original implementation fixed before unblinding. Path \pathB preserves hidden states read from block 19 and reconstructs $B_4$ using a stabilized tail in the same forward pass. The 8,664 H1 block-level hidden-state files are elementwise identical to the existing canonical repooling cache. Path \pathC removes $B_4$ from both arms to examine dependence on that block. The set and order of paths and their interpretation were fixed before accessing true labels. \pathA is the primary result; \pathB and \pathC report sensitivity to numerical handling and block dependence.

The one-time confirmatory analysis invokes both arms, all three paths, and all estimand branches in a prespecified order. Validity of each branch's estimate follows the stated criteria and audit records. The protocol SHA-256 begins with \code{d483914c}; the finalized-record hash begins with \code{38a20718}; and the analysis-input hash begins with \code{2bad2870}. Unblinding occurred at \code{2026-08-23T05:47:11.928-04:00}. The confirmatory result SHA-256 is \code{ba467c67\ldots9eeff5a}. Full hashes are retained in the analysis-input and result-lineage manifests. Analyses added after unblinding are identified and reported separately as non-confirmatory.

\subsection{Estimands and Group Definitions}
\label{app:estimands}

The primary estimand E1, entity generalization E2, vocabulary-block generalization E3, and label-dependent grouping M2c use the following split rules.

\begin{itemize}
  \item \textbf{E1 (primary estimand).} Outer splits keep prompt groups disjoint to estimate conditional predictive gain when training and test sets share no prompt groups. H1 has six generation records per prompt, all assigned to the same fold. Each MCQ question forms a separate group.
  \item \textbf{E2 (entity generalization).} H1 is split by connected components of the identifier co-occurrence graph so that an identifier never crosses folds. E2 is non-estimable if the largest component covers more than $20\%$ of records or there are fewer than $5F=25$ components; only graph structure is then reported. Here $F=5$ is the fixed number of outer folds, and 25 components ensure at least five per fold on average. These estimability conditions were prespecified for this grouping and cross-validation design.
  \item \textbf{E3 (vocabulary-block generalization).} H1's identifier vocabulary is partitioned into five fixed blocks without labels; the blocks serve as outer test folds. Labels use only primary identifier mentions within the current vocabulary block, and hidden states are pooled only over token positions corresponding to that block.
  \item \textbf{M2c (label-dependent grouping sensitivity).} H1 is split by connected components formed from prompts and fabricated identifiers. Because grouping uses outcome labels, M2c is reported as a grouping-sensitivity analysis.
\end{itemize}

E2, E3, and M2c apply only to H1. MCQ lacks the corresponding identifier-entity and vocabulary-block structures, so all three are marked not applicable for MCQ.

Conditional gain can vary with baseline information, numerical handling, and grouping. To examine these dependencies, we remove information blocks, reconstruct affected features, and change sample grouping. Identity and format controls examine task constructs \citep{hewitt2019,gururangan2018,poliak2018}; identifier and vocabulary grouping examine performance under different sharing conditions \citep{roberts2017,elangovan2021}. \Cref{tab:analysis-condition-map} summarizes these comparisons and their analytical roles.

\input{tables/table_analysis_condition_map}

\section{Full Results and Checks}
\label{app:results}

\subsection{Prediction Results for All Paths}
\label{app:e1-full}
\label{sec:numeric-results}

Qwen paths A, B, and C are confirmatory, with values from the August 23, 2026 confirmatory results. Llama paths A and C are non-confirmatory. \Cref{tab:app-e1} reports all Qwen estimates. The left panel of \cref{fig:e1-results} compares AUROC for $\Sstar$, $H$, and $\Sstar+H$; the right panel shows the corresponding gains and 95\% confidence intervals. Gray shading identifies path C, and a broken axis accommodates its MCQ gain. \Cref{tab:app-mc-error} reports interval widths and their Monte Carlo error bands.

\begin{papertable}
  \centering\scriptsize
  \caption{Confirmatory estimates for the six E1 branches.}
  \label{tab:app-e1}
  \begin{tabular}{llrrrrrlrl}
    \toprule
    Path & Arm & $n$ & $G$ & $\Sstar$ & $H$ & $\Sstar+H$ & $\deltah$ (95\% CI) & CI width & Precision \\
    \midrule
    \pathA & H1  & 8,664 & 1,508 & 0.9889 & 0.9438 & 0.9905 & $0.00157$ $[0.00056,0.00283]$ & 0.00226 & Adequate \\
    \pathA & MCQ & 1,750 & 1,750 & 0.8198 & 0.8292 & 0.8487 & $0.02893$ $[0.01894,0.03895]$ & 0.02001 & Adequate \\
    \pathB & H1  & 8,664 & 1,508 & 0.9854 & 0.9438 & 0.9882 & $0.00282$ $[0.00137,0.00456]$ & 0.00319 & Adequate \\
    \pathB & MCQ & 1,750 & 1,750 & 0.8219 & 0.8292 & 0.8492 & $0.02728$ $[0.01747,0.03717]$ & 0.01970 & Adequate \\
    \pathC & H1  & 8,664 & 1,508 & 0.9853 & 0.9438 & 0.9891 & $0.00381$ $[0.00197,0.00592]$ & 0.00394 & Adequate \\
    \pathC & MCQ & 1,750 & 1,750 & 0.5684 & 0.8292 & 0.8251 & $0.25672$ $[0.22699,0.28668]$ & 0.05969 & Insufficient \\
    \bottomrule
  \end{tabular}

  \vspace{0.35em}
  \begin{minipage}{0.96\linewidth}
    \footnotesize
    $n$ is the number of records, and $G$ is the number of groups resampled by the grouped bootstrap.
  \end{minipage}
\end{papertable}

\textbf{Effect of numerical handling.} Path \pathB reconstructs affected generation features while preserving hidden states to test whether numerical handling changes the gain. The resulting gains are $0.0028$ on H1 and $0.0273$ on MCQ, preserving path A's direction and magnitude. \Cref{tab:app-e1,fig:e1-results} show that H1 gains are on the order of thousandths and MCQ gains on the order of hundredths under paths A and B, with all intervals excluding zero. Predictive improvements on both tasks remain stable after this reconstruction.

The C/MCQ gain is substantially larger but accompanies a weaker combined baseline. Its interval width is $0.05969$, exceeding the prespecified limit of $0.04$.

\begin{paperfigure}
  \centering
  \figslot{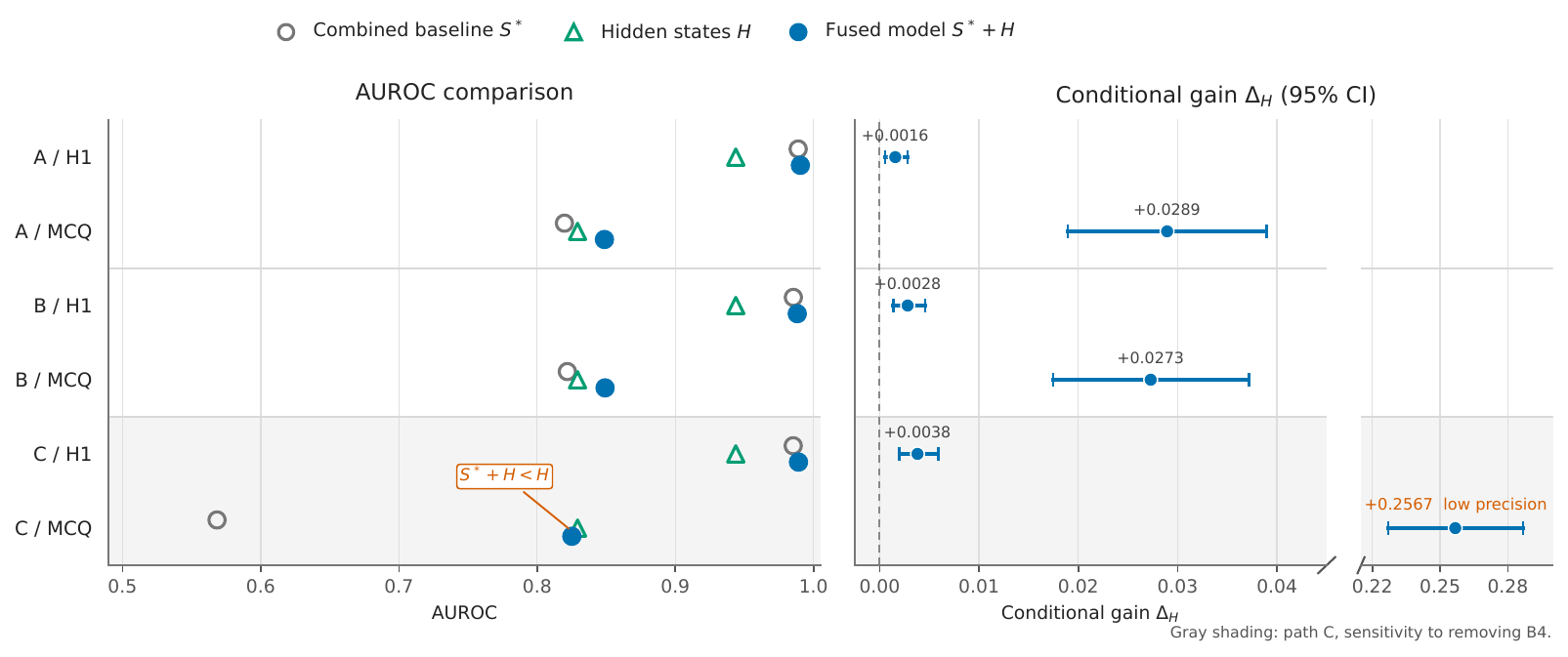}{0.40\textheight}
  \caption{AUROC and conditional gain across E1 paths.}
  \label{fig:e1-results}
\end{paperfigure}

\begin{papertable}
  \centering\small
  \caption{E1 confidence interval widths and Monte Carlo error bands.}
  \label{tab:app-mc-error}
  \begin{tabular}{llrrl}
    \toprule
    Path & Arm & CI width & Monte Carlo band for width & Precision \\
    \midrule
    \pathA & H1  & 0.00226 & $[0.00221,0.00232]$ & Adequate \\
    \pathA & MCQ & 0.02001 & $[0.01973,0.02028]$ & Adequate \\
    \pathB & H1  & 0.00319 & $[0.00312,0.00326]$ & Adequate \\
    \pathB & MCQ & 0.01970 & $[0.01951,0.01989]$ & Adequate \\
    \pathC & H1  & 0.00394 & $[0.00390,0.00399]$ & Adequate \\
    \pathC & MCQ & 0.05969 & $[0.05881,0.06058]$ & Insufficient \\
    \bottomrule
  \end{tabular}
\end{papertable}

Jackknife standard errors of the endpoints range from $7.6\times10^{-6}$ to $4.2\times10^{-4}$, indicating that 20,000 bootstrap samples reliably distinguish the precision classifications in this table. Monte Carlo error captures numerical variation from finite resampling; confidence intervals capture sample variation estimated through group resampling.

\Cref{tab:llama-results} reports Llama's paths A and C.

\input{tables/table_llama_results}

\subsection{Blockwise Performance and Baseline Comparisons}
\label{app:blockwise}
\label{sec:blockwise-results}

To assess preservation of individual blocks' predictive performance, \cref{tab:app-blocks} reports their AUROC values and the baseline's differences from them. These comparisons characterize H1's main predictive cues and MCQ's loss from combination.

\begin{papertable}
  \centering\small
  \caption{Single-block AUROC and baseline point-estimate differences on path \pathA.}
  \label{tab:app-blocks}
  \label{tab:block-results}
  \begin{tabular}{lrrrrrr}
    \toprule
    Arm and metric & $B_1$ & $B_2$ & $B_3$ & $B_4$ & $B_5$ & $\Sstar$ \\
    \midrule
    H1 \quad AUROC          & 0.9826 & 0.6662 & 0.9286 & 0.8023 & 0.6456 & 0.9889 \\
    H1 \quad $\Delta_{S,k}$ & 0.0063 & 0.3227 & 0.0603 & 0.1865 & 0.3433 & --- \\
    \midrule
    MCQ \quad AUROC          & 0.4884 & 0.5329 & 0.5304 & 0.8256 & 0.5783 & 0.8198 \\
    MCQ \quad $\Delta_{S,k}$ & 0.3314 & 0.2868 & 0.2894 & $-0.0059$ & 0.2415 & --- \\
    \bottomrule
  \end{tabular}
\end{papertable}

\textbf{Identifier identity versus text content.} To assess whether H1's high detection performance can already be explained by identifier identity, we compare $B_1$ with full-text block $B_3$. Their AUROC values in \cref{tab:app-blocks} are $0.9826$ and $0.9286$, respectively: identity alone performs better. Identifier identity therefore supplies strong error-prediction cues. The AUROC gap between $B_1$ and the combined baseline is $0.0063$.

On MCQ, $\Sstar$ falls 0.0059 below $B_4$. This difference is within the prespecified non-inferiority tolerance, and M1b passes. The value 0.0059 is a descriptive difference between point estimates; its confidence interval was not separately estimated. $\deltah=0.02893$ measures the total improvement over the five-block stacked baseline.

\subsection{Numerical Results for the Criteria}
\label{app:criteria-values}

This section reports criterion outcomes, the most constraining blocks, and lower confidence bounds for the six E1 branches. Direct blockwise improvement and the more conservative consistency requirement are distinguished, using frozen confirmatory results throughout. iv-b requires M1b to pass and the lower confidence bound for $\deltah$ to exceed $\varepsilon_{NI}=0.02$.

\begin{papertable}
  \centering\small
  \caption{Blockwise criteria for the six E1 branches.}
  \label{tab:app-e1-criteria}
  \begin{tabular}{llllll}
    \toprule
    Path & Arm & iv-a min. bound (block) & M1b & $\deltah$ lower bound & iv-b \\
    \midrule
    \pathA & H1  & $0.0048$ ($B_1$) & Pass & $0.0006$ & Fail \\
    \pathA & MCQ & $0.0145$ ($B_4$) & Pass & $0.0189$ & Fail \\
    \pathB & H1  & $0.0016$ ($B_1$) & Pass & $0.0014$ & Fail \\
    \pathB & MCQ & $0.0158$ ($B_4$) & Pass & $0.0175$ & Fail \\
    \pathC & H1  & $0.0038$ ($B_1$) & Pass & $0.0020$ & Fail \\
    \pathC & MCQ & $0.2194$ ($B_5$) & Fail ($B_5$) & $0.2270$ & Fail \\
    \bottomrule
  \end{tabular}

  \vspace{0.35em}
  \begin{minipage}{0.96\linewidth}
    \footnotesize
    The iv-a column gives the smallest marginal one-sided 95\% lower bound for the fused model's difference from individual block predictors.
  \end{minipage}
\end{papertable}

In \cref{tab:app-e1-criteria}, paths A and B and C/H1 fail iv-b because their gain lower bounds do not exceed $0.02$. Although C/MCQ has a lower bound of $0.2270$, the baseline fails non-inferiority against $B_5$, so it also fails the conjunction required by iv-b.

For both tasks on Qwen path A, the fused model's differences from all individual blocks have positive lower confidence bounds, supporting superiority over every compared block. The following path-\pathA\ bounds show the M1b and iv-a calculations, with thresholds of $-0.02$ and $0$, respectively.

\begin{papertable}
  \centering\small
  \caption{Marginal one-sided 95\% lower bounds for M1b and iv-a on path \pathA.}
  \label{tab:app-bounds}
  \begin{tabular}{llrrrrrl}
    \toprule
    Criterion & Arm & $B_1$ & $B_2$ & $B_3$ & $B_4$ & $B_5$ & Decision \\
    \midrule
    M1b  & H1  & 0.0036 & 0.2969 & 0.0500 & 0.1653 & 0.3188 & Pass \\
    M1b  & MCQ & 0.3007 & 0.2572 & 0.2603 & $-0.0112$ & 0.2131 & Pass \\
    iv-a & H1  & 0.0048 & 0.2987 & 0.0514 & 0.1667 & 0.3204 & Pass \\
    iv-a & MCQ & 0.3310 & 0.2866 & 0.2912 & 0.0145 & 0.2430 & Pass \\
    \bottomrule
  \end{tabular}
\end{papertable}

\Cref{fig:block-bounds} subtracts each criterion's threshold from the marginal one-sided 95\% lower bounds on the primary path. The non-inferiority panels show lower bounds plus $0.02$; the direct-comparison panels show the lower bounds themselves. All points lie to the right of zero, so every corresponding blockwise check passes. Blocks closer to zero constrain the joint decision more strongly.

\begin{paperfigure}
  \centering
  \includegraphics[width=\linewidth]{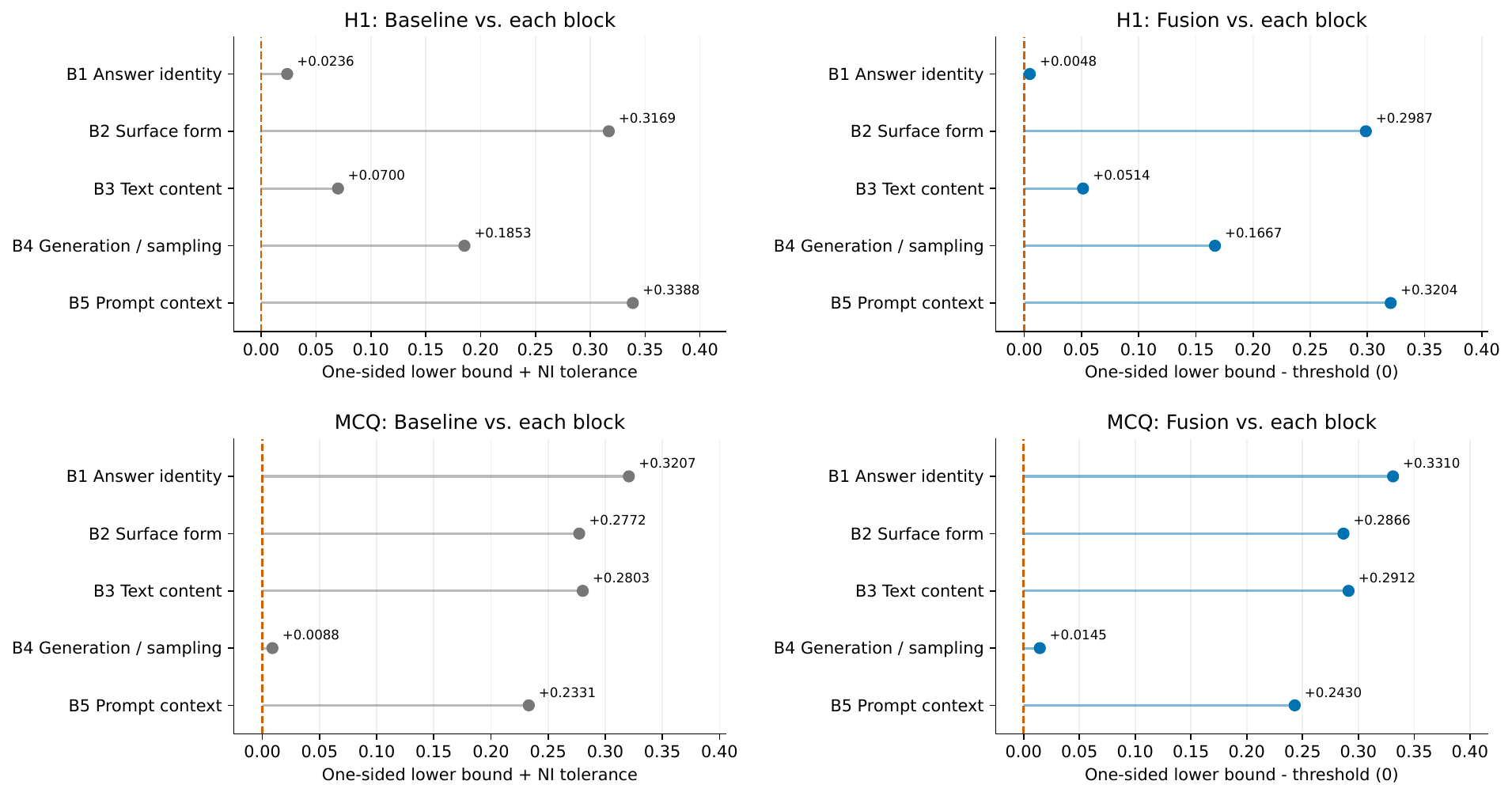}
  \caption{Distance between blockwise lower bounds and criterion thresholds on primary path A. Points show the marginal one-sided 95\% lower bound minus its threshold; positive values satisfy the corresponding criterion.}
  \label{fig:block-bounds}
\end{paperfigure}

On path \pathA, both arms fail iv-b. The MCQ lower bound for $\deltah$ is 0.018941, falling 0.00106 below 0.02. H1's bound is 0.000562, an order of magnitude below the threshold. Both arms pass the direct blockwise comparison iv-a but fail the more conservative conjunction in iv-b.

The path-\pathA\ task-construct check is triggered on H1 by $B_1$: $\Delta_{\mathrm{task},1}=-0.0539$, with marginal one-sided upper bound $-0.0437$. The $B_2$ difference is 0.2624 and does not trigger the check. Neither MCQ comparison triggers it. H1 thus shows strong prediction from identity cues under the current label definition; conditional gain is still reported under the same evaluation protocol.

Both Llama tasks on path A pass M1b and iv-a. H1's smallest non-inferiority and direct-comparison lower bounds are $0.0017$ and $0.0057$, both constrained by $B_1$. MCQ's corresponding bounds are $-0.0165$ and $0.0372$, both constrained by $B_4$ and exceeding $-0.02$ and zero, respectively. H1's gain lower bound is $0.0014$, so iv-b fails; MCQ's is $0.0447$, so it passes. With joint option-text and full-response encoding, Llama MCQ still passes M1b and iv-a, and its gain lower bound of $0.0264$ also passes iv-b.

\subsection{Additional Estimands and Incomplete Analyses}
\label{app:other-estimands}
\label{sec:generalization-results}

We report generalization and grouping-sensitivity results, followed by the reasons why the cross-arm contrast and RQ2 did not yield valid estimates.

\subsubsection{Generalization and Grouping Sensitivity}

To examine generalization to unseen identifiers, E2 connects records sharing a prompt or identifier and uses connected components to define disjoint training and test groups. H1's entity co-occurrence graph contains 889 identifier identities and 10,494 identity edges, forming 17 components, fewer than the required 25. The largest component covers 8,595/8,664 records ($99.20\%$), exceeding the $20\%$ limit. Both prespecified structural conditions fail, so E2 is non-estimable and the analysis stops at structural statistics.

E3 assigns 1,445 prompt groups to five fixed vocabulary blocks. No target block can be assigned to 161 records, and another 2,418 records lack eligible labels and hidden states within their own test block, leaving 6,085 analyzable records. All five test folds and their training complements contain both classes.

\begin{papertable}
  \centering\scriptsize
  \caption{Confirmatory E3 estimates and criteria.}
  \label{tab:app-e3}
  \begin{tabular}{lrrrrlll}
    \toprule
    Path & $\Sstar$ & $H$ & $\Sstar+H$ & $\deltah$ (95\% CI) & M1b & iv-a & iv-b \\
    \midrule
    \pathA & 0.8094 & 0.8319 & 0.8379 & $0.0285$ $[0.0142,0.0434]$ & Not certified ($B_4$) & Pass & Fail \\
    \pathB & 0.8114 & 0.8319 & 0.8382 & $0.0268$ $[0.0116,0.0426]$ & Certified & Pass & Fail \\
    \pathC & 0.7629 & 0.8319 & 0.8192 & $0.0562$ $[0.0382,0.0747]$ & Certified & Pass & Pass \\
    \bottomrule
  \end{tabular}
\end{papertable}

In \cref{tab:app-e3}, all three E3 gain intervals exclude zero and all pass direct comparison iv-a, but the remaining criteria differ. Path A fails baseline non-inferiority against $B_4$. Path B passes non-inferiority, but its gain lower bound remains below $0.02$. Only path C satisfies both conditions and passes iv-b. E3 estimates conditional gain after redefining labels and hidden states by vocabulary block.

M2c uses 238 fabricated identifiers and 1,124 identity edges, producing 1,283 components with a largest-component share of 0.0464. Gains across the three paths are 0.0142, 0.0205, and 0.0236, respectively, but all fail M1b against $B_1$ and all fail iv-a. This grouping prevents only fabricated identifiers from crossing folds; known correct identifiers may still be shared. Grouping also depends on labels, with an unknown direction of bias. These gain estimates characterize sensitivity to label-dependent grouping. Superiority of the fused model over every compared block is sensitive to the grouping rule.

\Cref{tab:other-estimands} reports the E3 and M2c criterion bounds and thresholds, identifying the blocks or conditions that constrain these decisions.

\input{tables/table_03_other_estimands}

\subsubsection{Cross-Arm Contrast and Stratified Analysis}

The prespecified cross-arm contrast is $D=S_{MCQ}-S_{H1}$ and requires out-of-fold predictions from the output-side observable predictor $O$, which combines generation scores and visible output text. The implementation mistakenly supplied out-of-fold predictions from the fused model $\Sstar+H$, so the prespecified $D$ was not validly estimated. The frozen procedure returned nominal estimates of $-0.2835$, $-0.2781$, and $-0.3280$, with nominal 95\% intervals $[-0.3262,-0.2413]$, $[-0.3210,-0.2351]$, and $[-0.3731,-0.2836]$, respectively. These values are retained only as audit records of the input mismatch and cannot determine the prespecified cross-arm gate or compare the tasks. Recalculation under the original definition would be a post hoc recovery analysis.

RQ2 was intended to stratify by identifier-token log-probability and describe conditional gain at different observable confidence levels. Its prespecified primary stratifier depends on the H1 $B_4$ field affected by non-finite-value filtering. Following the branch fixed before unblinding, the primary analysis marks it uninterpretable and stops computation. The stratifier was not replaced with tail-stabilized fields, so no confirmatory RQ2 estimates are reported.

\section{Discussion and Limitations}
\label{app:interpretation}

\input{sections/07_discussion}

\input{sections/09_limitations}

%% file: tables/table_01_population.tex

\begin{papertable}
  \centering\small
  \caption{Sample accounting for Qwen.}
  \label{tab:population}
  \begin{tabular}{llrrrrr}
    \toprule
    Arm & Unit & Final & Analyzable & Ineligible & Groups $G$ & Errors (rate) \\
    \midrule
    H1  & Generation & 9,264 & 8,664 & 600 & 1,508 & 693 (0.0800) \\
    MCQ & Question     & 1,751 & 1,750 & 1   & 1,750 & 318 (0.1817) \\
    \bottomrule
  \end{tabular}

  \vspace{0.35em}
  \begin{minipage}{0.96\linewidth}
    \footnotesize
    $G$ is the number of resampling groups: prompts for H1 and questions for MCQ. In H1, 345 prompt groups contain at least one error.
  \end{minipage}
\end{papertable}

%% file: tables/table_analysis_condition_map.tex
\begin{papertable}
  \centering
  \caption{Purposes and settings of additional comparisons.}
  \label{tab:analysis-condition-map}
  \small
  \setlength{\tabcolsep}{4pt}
  \begin{tabularx}{\textwidth}{@{}>{\raggedright\arraybackslash}p{0.22\textwidth}>{\raggedright\arraybackslash}X>{\raggedright\arraybackslash}p{0.16\textwidth}@{}}
    \toprule
    Purpose & Comparison setting & Analysis role \\
    \midrule
    Dependence on generation and sampling signals & Path \pathC removes $B_4$ from the baseline and fused model in both arms & Block-dependence diagnostic \\
    Sensitivity to numerical handling & Path \pathB retains information coverage and hidden states, reconstructing affected features through tail stabilization for comparison with path \pathA & Protocol-deviation sensitivity \\
    Performance without shared identifiers (E2) & Split by prompt--identifier co-occurrence components without using error labels; do not estimate performance if structural conditions fail & Prespecified generalization evaluation \\
    Performance under changed vocabulary scope (E3) & Partition vocabulary by a fixed hash rule and redefine block-level labels and hidden states, yielding a separate estimand & Prespecified additional estimand \\
    Sensitivity to grouping (M2c) & Construct label-dependent groups from prompts and fabricated identifiers; use only for sensitivity because bias direction is unknown & Grouping sensitivity \\
    Dependence on identity and format cues & Compare full-text block $B_3$ with identity block $B_1$ and format block $B_2$ to check the task construct & Prespecified construct check \\
    \bottomrule
  \end{tabularx}
\end{papertable}

%% file: tables/table_llama_results.tex
\begin{papertable}
  \centering\small
  \setlength{\tabcolsep}{4pt}
  \caption{Llama detection performance and conditional gains on paths A and C.}
  \label{tab:llama-results}
  \begin{tabular}{lrrrl}
    \toprule
    Path / task & $\Sstar$ & $H$ & $\Sstar+H$ & $\deltah$ (95\% CI) \\
    \midrule
    A / H1 & 0.9712 & 0.9148 & 0.9752 & $+0.0040$ $[+0.0014,+0.0075]$ \\
    A / MCQ & 0.7465 & 0.7916 & 0.8061 & $+0.0596$ $[+0.0447,+0.0751]$ \\
    C / H1 & 0.9668 & 0.9148 & 0.9725 & $+0.0057$ $[+0.0021,+0.0104]$ \\
    C / MCQ & 0.5595 & 0.7916 & 0.7816 & $+0.2221$ $[+0.1937,+0.2506]$ \\
    \bottomrule
  \end{tabular}
  \vspace{0.3em}
  \begin{minipage}{0.97\linewidth}
    \footnotesize The first three numerical columns report AUROC. H1 has 8,810 records and 1,516 groups; MCQ has 1,746 records and 1,746 groups. A retains all five blocks; C removes $B_4$. C/MCQ fails baseline non-inferiority and has insufficient interval precision.
  \end{minipage}
\end{papertable}

%% file: tables/table_03_other_estimands.tex

\begin{papertable}
  \centering
  \caption{Vocabulary-block generalization (E3) and grouping sensitivity (M2c).}
  \label{tab:other-estimands}
  \resizebox{\textwidth}{!}{%
  \begin{tabular}{llrlccc}
    \toprule
    Estimand & Path & $n$ & $\deltah$ (95\% CI) & \shortstack{Non-inferiority\\Min. bound / threshold} & \shortstack{Direct comparison\\Min. bound / threshold} & \shortstack{Consistency\\$\deltah$ bound / threshold} \\
    \midrule
    E3 & \pathA & 6,085 & $+0.0285$ $[0.0142,0.0434]$ & $-0.0229/-0.0200$ ($B_4$) & $0.0053/0$ ($B_4$) & $0.0142/0.0200$ \\
    E3 & \pathB & 6,085 & $+0.0268$ $[0.0116,0.0426]$ & $-0.0122/-0.0200$ ($B_4$) & $0.0139/0$ ($B_4$) & $0.0116/0.0200$ \\
    E3 & \pathC & 6,085 & $+0.0562$ $[0.0382,0.0747]$ & $0.0065/-0.0200$ ($B_1$) & $0.0519/0$ ($B_1$) & $0.0382/0.0200$ \\
    \midrule
    M2c & \pathA & 8,664 & $+0.0142$ $[0.0087,0.0192]$ & $-0.0463/-0.0200$ ($B_1$) & $-0.0332/0$ ($B_1$) & $0.0087/0.0200$ \\
    M2c & \pathB & 8,664 & $+0.0205$ $[0.0136,0.0270]$ & $-0.0564/-0.0200$ ($B_1$) & $-0.0385/0$ ($B_1$) & $0.0136/0.0200$ \\
    M2c & \pathC & 8,664 & $+0.0236$ $[0.0177,0.0291]$ & $-0.0637/-0.0200$ ($B_1$) & $-0.0415/0$ ($B_1$) & $0.0177/0.0200$ \\
    \bottomrule
  \end{tabular}}

  \vspace{0.35em}
  \begin{minipage}{0.98\linewidth}
    \footnotesize
    $n$ is the record count. Criterion columns give lower bound / threshold; parentheses identify the block with the smallest bound. Consistency requires non-inferiority and a gain lower bound above $0.02$.
  \end{minipage}
\end{papertable}

%% file: sections/07_discussion.tex
\subsection{Additional Discussion}
\label{app:discussion}

Output-baseline performance, task labels, and sample grouping jointly shape the interpretation of hidden-state gains.

Hidden states and output-side information may share predictive cues. Conditional gain measures the additional predictive capability obtained by the current learning procedure beyond existing output information. H1's output baseline approaches the AUROC ceiling, limiting absolute headroom. Qwen's baseline is $0.9889$: even a fused AUROC of $1$ would leave only about $0.0111$ of improvement at the point-estimate level. Llama H1 and the tested Qwen readout adaptations likewise show small gains under the current labels, learners, and evaluation populations.

MCQ retains more pronounced gains. Llama improves by $0.0387$ after joint full-response encoding, showing that positive gains persist with broader output-text coverage. The same result illustrates coverage dependence: adding the full response changes both baseline and fused performance and reduces the gain estimate.

Blockwise checks define the baseline's comparison scope. Non-inferiority assesses the combined baseline against the selected blocks within a specified tolerance. Passing direct comparisons supports fused-model superiority over those blocks; consistency is a separate, more conservative condition. All six Qwen E1 branches fail the consistency check, whereas Llama MCQ path A and the joint full-response condition pass. The lower-bound threshold for this study's consistency check is $0.02$.

Reducing baseline information can increase the gain while decreasing final detection performance. This occurs on MCQ path C for both models. Ranking readout methods solely by gain would therefore overlook changes in the reference baseline. Studies should report the combined baseline, standalone hidden-state performance, fused performance, and conditional gain together, specifying the output sources and blockwise baseline comparisons.

Future output-side comparisons could include semantic entropy, SelfCheckGPT, and other text encoders, separately estimating changes in baseline and fused performance. The effect of the parsing-failure denominator also requires re-estimating both.

H1 labels depend on identifier existence in a fixed knowledge base, directly linking identity to the label. Identity alone supplies strong error-prediction cues. In standalone-statement adaptation, about $93.4\%$--$93.8\%$ of test-record texts occur in the corresponding training fold. Identifier memorization is one possible explanation, but its contribution has not been separated from generalization. E2 is non-estimable, leaving generalization to unseen identifiers without a valid estimate. E3 concerns a prediction task with labels and hidden states redefined by vocabulary block. M2c retains positive gains but fails direct blockwise comparison, further limiting generalization across grouping schemes.

%% file: sections/09_limitations.tex
\subsection{Limitations and Unevaluated Comparisons}
\label{sec:limitations}

The experiments leave the following conditions and data-processing limitations unresolved:
\begin{enumerate}
  \item \textbf{Models, tasks, and analysis status.}
  We study Qwen2.5-7B-Instruct and Llama-3.1-8B-Instruct. Qwen A/B/C analyses are confirmatory; Llama A/C analyses are non-confirmatory. Both use local 8-bit quantized backends; other precisions and domains are not covered. Gains are estimated under each model's generation budget, parser version, and analyzable population. H1 labels verify whether generated identifiers exist in a fixed knowledge base.

  \item \textbf{Information access and output baselines.}
  The study uses decoding scores and repeated-sampling statistics, requiring access to those signals. Output controls cover five block types and TF-IDF text encoding; other output-side methods and encoders remain to be compared. Blockwise non-inferiority applies to the selected blocks and prespecified tolerance.

  \item \textbf{Readout settings and adaptation scope.}
  The main analysis uses a fixed layer and linear readout, while additional experiments cover limited MLP, L1, and XGBoost settings. MLP candidate training budgets are limited, and probability clipping in the earlier readout conditions may change extreme scores and fusion inputs. Actual iteration counts for selected models and complete execution traces were not retained. H1 representation adaptation still has upper-bound tuning selections, extensive cross-fold text duplication, and missing intermediate logs. Results concern these representations, budgets, and scoring settings.

  \item \textbf{Numerical integrity and parsing conventions.}
  Non-finite-value filtering in Qwen's original H1 $B_4$ continues to constrain related conclusions. Llama's $B_4$ has not undergone an independent numerical integrity audit. Although all target-layer vectors in Qwen statement extraction are finite, downstream layers and output scores widely contain non-finite values. MCQ consistency features use successful parses as the denominator, affecting 128/1,746 Llama analysis rows through failure exclusion. Recalculation agrees with the current implementation, but the failure-denominator rule was not specified in the original protocol. Its directional effect on predictive performance and gain has not been measured.

  \item \textbf{Absolute gain and practical utility.}
  A baseline near the AUROC ceiling limits absolute headroom. Label composition and baseline levels across tasks and models also affect comparisons. This study evaluates ranking performance through AUROC gain and uses an operational tolerance of $0.02$. Decisions at fixed thresholds, probability calibration, and readout cost remain unevaluated.
\end{enumerate}